\documentclass[acmtog]{acmart}

\AtBeginDocument{%
  }

\acmJournal{TOG}
\acmVolume{37}
\acmNumber{4}
\acmArticle{111}
\acmMonth{8}

\begin{document}

\title{Surface Reconstruction from Point Clouds via Grid-based Intersection Prediction}

\author{Hui Tian}
\affiliation{%
  \institution{School of Computer, National University of Defense Technology}
  \streetaddress{Deya Road 109 Changsha, Hunan, China}
  \city{Changsha}
  \country{China}}
\email{tianhui13@nudt.edu.cn}

\author{Kai Xu}
\affiliation{%
  \institution{School of Computer, National University of Defense Technology}
  \streetaddress{Deya Road 109 Changsha, Hunan, China}
  \city{Changsha}
  \country{China}}
\email{kevin.kai.xu@gmail.com}

\renewcommand{\shortauthors}{Hui Tian and Kai Xu}

\begin{teaserfigure}
\centering
\includegraphics[width=1.0\textwidth]{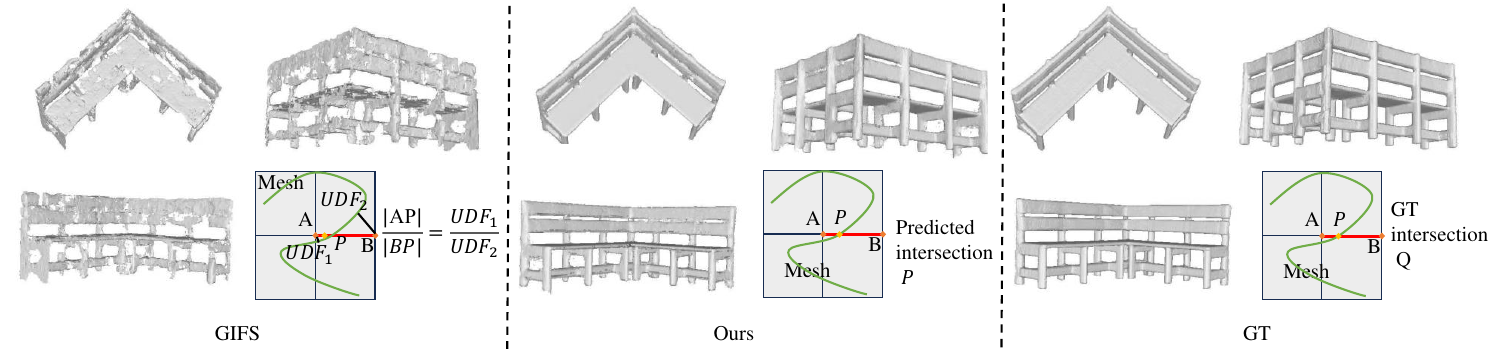}
\caption{The illustration and mesh of GIFS~\cite{gifs}, Ours and GT. For every gird, the picture in the right down corner is the illustration of this method, others are 3 views of the shape. $P$ is the intersection of cube edge and surface. $A$ and $B$ are the vertices at the corner of the cube. We can see that our reconstructed surface is with better details and less artefacts. The number of point in input point cloud is $3000$.}
\label{fig:teaser}
\end{teaserfigure}

\begin{abstract}
Surface reconstruction from point clouds is a crucial task in the fields of computer vision and computer graphics. SDF-based methods excel at reconstructing smooth meshes with minimal error and artefacts but struggle with representing open surfaces. On the other hand, UDF-based methods can effectively represent open surfaces but often introduce noise, leading to artefacts in the mesh. In this work, we propose a novel approach that directly predicts the intersection points between line segment of point pairs and implicit surfaces. To achieve it, we propose two modules named Relative Intersection Module and Sign Module respectively with the feature of point pair as input. To preserve the continuity of the surface, we also integrate symmetry into the two modules, which means the position of predicted intersection will not change even if the input order of the point pair changes. This method not only preserves the ability to represent open surfaces but also eliminates most artefacts on the mesh. Our approach demonstrates state-of-the-art performance on three datasets: ShapeNet, MGN, and ScanNet. The code will be made available upon acceptance.
\end{abstract}



\received{20 February 2007}
\received[revised]{12 March 2009}
\received[accepted]{5 June 2009}


\maketitle

\section{Introduction}
Surface reconstruction from point clouds is a long-standing and essential task that has been studied for many years. It plays a crucial role in various modern applications, including visual navigation and robotics. Numerous notable works~\cite{2022dynamicCode, deepmls, 2018Deepmc, 2020ndf, 2020npull, SAL, SALD, dude, dualoctree, uwed, 2020SSRNet, Variational} have significantly advanced this field. Among the traditional methods, Poisson Surface Reconstruction~\cite{2013Poisson}, a well-known approach, has yielded good results. However, its performance may deteriorate in the absence of normals and faces with open-surface challenges.
With the emergence of deep learning, many researchers have sought to enhance reconstruction results by leveraging the robust representation and predictive capabilities of deep neural networks. These methods can generally be categorized into SDF-based (Signed Distance Function) and UDF-based (Unsigned Distance Function) approaches in terms of geometry representation. UDF-based methods excel at representing open surfaces, whereas SDF-based methods struggle in this regard. Yet, predicting UDF values often introduces noise, despite the critical role these UDF values play in achieving high-quality surface reconstruction. In this study, a novel approach is proposed to enhance surface reconstruction by directly predicting the intersection points between surfaces and line segments. This method not only improves reconstruction quality but also maintains the capacity to represent open surfaces effectively.

\begin{figure} 
\centering
  \includegraphics[width=\columnwidth]{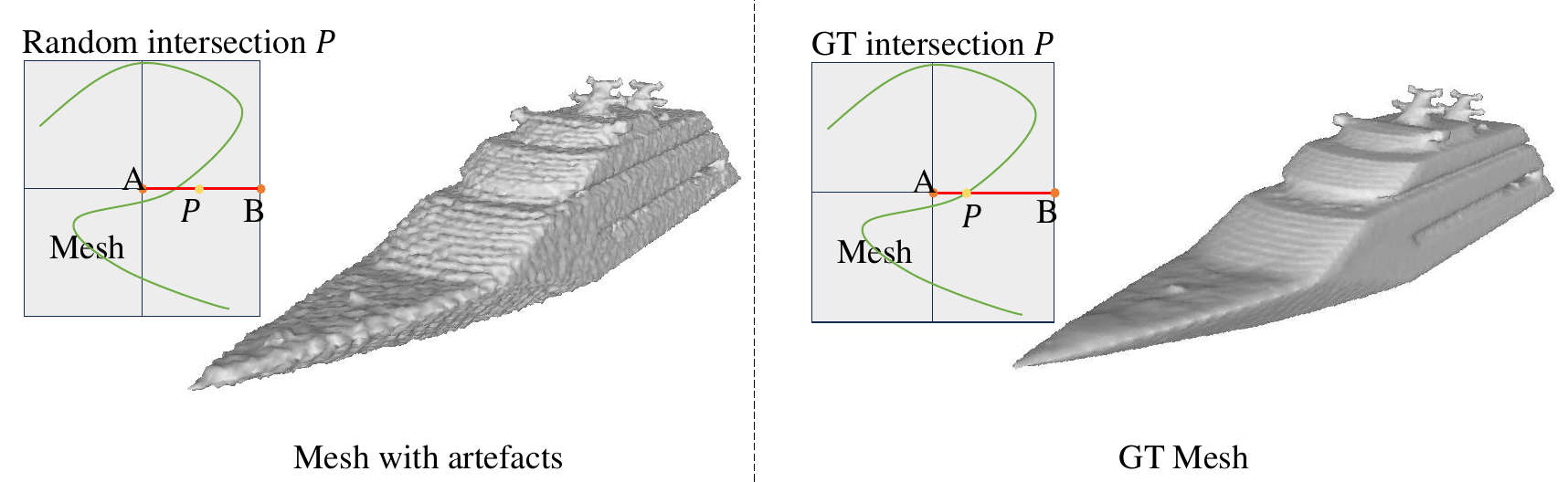}
  \caption{The left column is shape visualization when we keep the vertices sign at the corner of cube correct and give the intersection a random disturbing. The right column is the GT shape.}
  \label{fig:artefacts}
\end{figure}

\begin{figure} 
\centering
  \includegraphics[width=0.75\columnwidth]{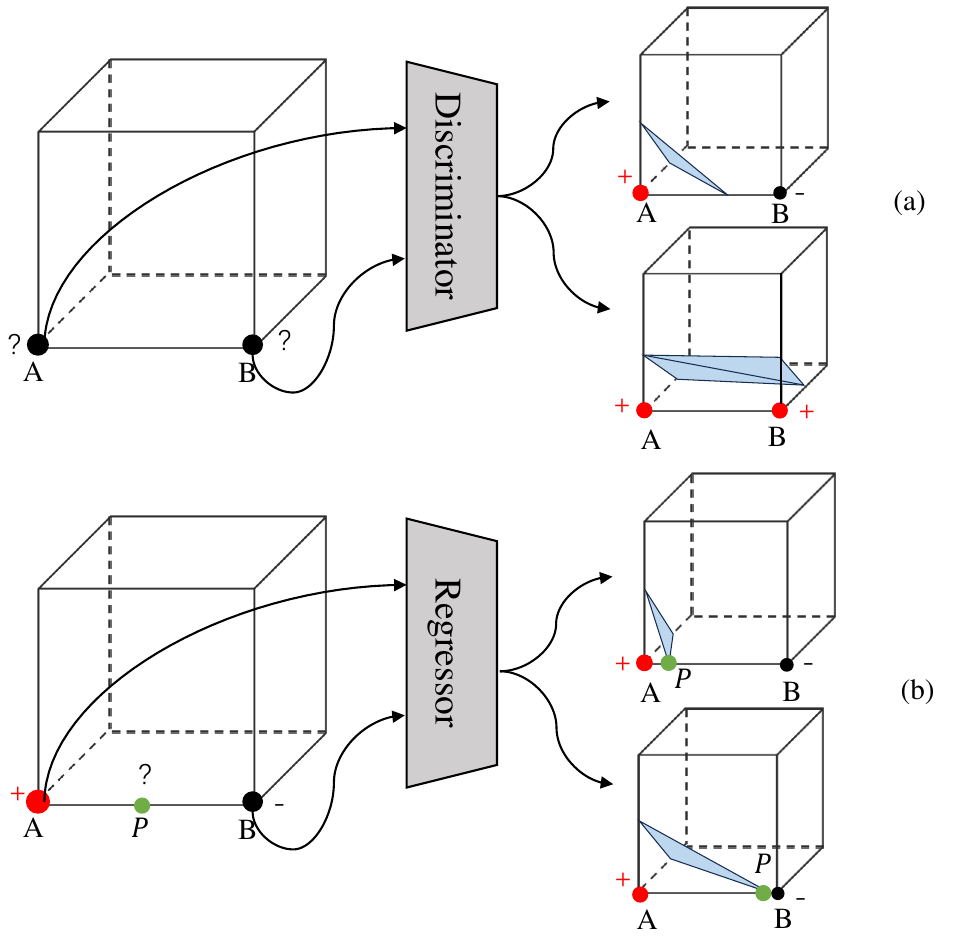}
  \caption{This figure indicates the two parts in reconstructing mesh, the sign of vertices of 8 corners in a cube in (a) and the intersection position in (b).}
  \label{fig:cube}
\end{figure}

In particular, the motivation of our method comes from the following two aspects. First, Most methods for reconstructing surfaces from point clouds ultimately rely on Marching Cubes or its variants. Marching Cubes requires two key pieces of information to generate a triangle mesh. As depicted in Fig.~\ref{fig:cube} (a), the first piece is the sign of the vertices at the eight corners of the cube, while the second is the intersection points between the surface and the edges of the cube, as the exact position of point P after determining that P lies on line AB in Fig.~\ref{fig:cube} (b). Therefore, in order to achieve better surface reconstruction, it is crucial to enhance the accuracy of both the sign of the vertices at the eight corners of a cube and the exact intersection positions. While most researchers agree on the importance of the accuracy of the sign of the vertices at the eight corners of a cube, the significance of the exact intersection position is often overlooked. However, the second aspect is also very important. This can be illustrated through a simple comparison. As demonstrated in Fig.~\ref{fig:artefacts}, the shape in the left column is reconstructed using Marching Cubes with ground-truth signs, and all intersection points are randomly chosen in $(0, 1)$, supposing the length of the cube edge is $1$. On the other hand, the shape in the right column is reconstructed with ground-truth signs and ground-truth intersection positions. It is evident that the shape in the right column appears smoother, whereas the one in the left column exhibits numerous artefacts. These artefacts mainly stem from incorrect intersection positions when $\alpha$ is disturbed by the random noise, where $\alpha=\frac{|AP|}{|AB|}$ in Fig.~\ref{fig:teaser}. Therefore, the accuracy of intersection positions significantly impacts the quality of surface reconstruction.

Secondly, it is a common phenomenon that predicting the UDF in the vicinity of the surface tends to introduce noise. This noise arises from the discontinuity of the UDF gradient on the two side of the surface. When utilizing a neural network to represent the UDF, accurately predicting the sharp changes in the UDF gradient around the zero-level set becomes a challenging task. Furthermore, we provide a small yet compelling example to illustrate this point in Section~\ref{sec:hard_udf}. The CAP method~\cite{capudf} also emphasizes this challenge. The presence of UDF noise can result in inaccurate intersection points between the cube edge and the surface, leading to artefacts in the reconstructed mesh. This discrepancy elucidates why SDF-based methods typically produce smoother results compared to UDF-based approaches. Building upon these insights, we aim to propose a method that can not only reconstruct the open surface but also mitigate the impact of UDF noise.

Therefore, we have chosen to avoid predicting UDF and instead focus on directly predicting the intersection points between surfaces and cube edges. Specifically, we have designed two modules: the Relative Sign Module and the Intersection Module.

In the Relative Sign module, we predict whether two points have the same sign and incorporate a symmetrical design in the module. The symmetry guarantees that the relative sign will not change when the input order of point pair changes. Once we confirm that the signs of the two points are different, we proceed to use the Intersection Module.

In the Intersection Module, we directly predict the intersection points between the cube edges and the surface, while also implementing a symmetrical design. The symmetry aims to achieve the following purpose. In the point pair with A as start point and B as end point, the intersection point will not change when A and B exchange their position. The symmetry of both the Intersection Module and the Relative Sign module ensures the continuity of the surface.

To demonstrate the superiority of our approach, we conducted evaluations using chamfer distance and normal consistency metrics across three diverse datasets: ShapeNet, MGN, and ScanNet, encompassing watertight shapes, open surfaces, and partially scanned scenes. Compared to state-of-the-art methods, our method consistently outperforms them, yielding the best results across all three datasets. Notably, our method excels on ShapeNet, demonstrating a remarkable 30\% reduction of $CD_1$ and a 1.8\% enhancement of $NC$.


In summary, we can list our main innovative points,
\begin{itemize}
    \item Initially proposed to directly predict the intersection point between the cube edge and the surface, leading to enhanced accuracy in determining the intersection point.
    \item Developed two specialized modules designed to predict these intersections while maintaining the surface's continuity.
    
    \item Successfully achieved high-quality reconstruction results on three datasets: ShapeNet, MGN, and ScanNet.
\end{itemize}

\section{Related Works}

\paragraph{Traditional point cloud reconstruction.} Point cloud reconstruction has been a persistent challenge in the fields of graphics and computer vision. Among the traditional methods, Poisson surface reconstruction~\cite{2013Poisson} and ball-pivoting reconstruction~\cite{ballpivoting} stand out as the most significant. The Poisson surface reconstruction method categorizes the query points based on the Poisson indicator function, while the ball-pivoting method generates a continuous surface by simulating a ball rolling over the points. Although these approaches yield satisfactory surface reconstructions, there is still room for further enhancement in performance.

\paragraph{SDF-based implicit surface reconstruction.} 
Implicit surface reconstruction methods based on signed distance functions (SDF) utilize deep learning techniques to either classify the occupancy of query points or directly predict the SDF value. These methods can be categorized into global approaches, which leverage overall shape information to classify query points, and local approaches, which classify query points based on their neighboring points.

Early representative works include AtlasNet\cite{atlasnet}, IF-Net\cite{ifnet}, DeepSDF\cite{2019DeepSDF} and BSP-Net\cite{chen2020bsp}, which extract shape features and reconstruct the surface from these features. Later, ConvOccNet\cite{2020Convolutional}, SSR-Net\cite{2020SSRNet}, DeepMLS\cite{deepmls, imls}, and POCO\cite{2022POCO}, O-CNN\cite{o-cnn} DeepCurrents\cite{deepcurrents}, LP-DIF\cite{lpdif} and Venkatesh\cite{venkatesh2021deep} develop this methods. ConvOccNet\cite{2020Convolutional} converts point cloud features into voxels and enhances features through volume convolution. SSR-Net\cite{2020SSRNet} extracts point features, maps neighborhood point features to octants, and classifies them accordingly. DeepMLS\cite{deepmls, imls} predicts normal and radius for each point and classifies query points based on the moving least-squares equation. ALTO\cite{alto, hu2020jittor} dynamically adjusts the representation of local geometry. Dual Octree\cite{dualoctree} proposes a new graph convolution operator defined over a regular grid of features
fused from irregular neighboring octree nodes at different levels, simultaneously improving the performance and reducing the cost. Li, T\cite{2022dynamicCode} propose a dynamic code to represent the local geometry with dynamic resource adaptively. Points2Surf\cite{2020Points2Surf} aims to regress the absolute SDF value using local information and determine the sign using global information.

Other SDF-based implicit surface reconstruction methods include SAL\cite{SAL}, SALD\cite{SALD}, StEiK\cite{yang2023steik}, SIREN\cite{siren}, DiGS\cite{digs} and On-Surface Prior\cite{onprior}, which aim to transform explicit representations like point clouds and 3D triangle soups into implicit SDF representations. These methods require unique training processes and network parameters for each 3D model. SAL\cite{SAL} employs MLP to predict shape SDFs with UDF-based metric supervision, while SALD\cite{SALD} adds a derivative regularization term. On-Surface Prior\cite{onprior} utilizes a pre-trained UDF-based network to improve SDF predictions.

Differential Marching Cube methods is related to ours, including Deep Marching Cube\cite{2018Deepmc} and Neural Marching Cube\cite{neuralmc}. Both of them are based on SDF.

While SDF methods have shown significant advancements in point cloud reconstruction, they have limitations in representing open surfaces or partial scans. This is important in realistic application, because we usually cannot guarantee that the scan is watertight.
\paragraph{UDF-based implicit surface reconstruction.}  
UDF has been a focal point in the realm of surface representation due to its ability to capture more general surfaces, including open surfaces and partial scans. Various research efforts have delved into leveraging UDF for implicit surface reconstruction. For instance, NDF~\cite{2020ndf} utilizes UDF to encode surface information and introduces a method for up-sampling point clouds during mesh reconstruction, employing the Ball Pivoting~\cite{ballpivoting} technique. On the other hand, DUDE~\cite{dude} adopts a volumetric UDF representation encoded within a neural network to describe shapes, leading to promising results in implicit surface reconstruction tasks. Nevertheless, a common challenge across these approaches is the reliance on 3D ground truth labels for training. Furthermore, a key unresolved issue pertains to the direct extraction of iso-surfaces from UDF. In a different approach, UWED~\cite{uwed} leverages Moving Least Squares (MLS) to convert UDF into a dense point cloud, showcasing an alternative strategy for handling UDF representations in surface reconstruction tasks. MeshUDF\cite{2021MeshUDF}vote the sign of local cube corner according the local geometry. HSDF\cite{Wang22HSDF} adopts the hybrid SDF and UDF to represent open surface.


\section{Methods}

\subsection{Overview}
\begin{figure*} 
\centering
  \includegraphics[width=2\columnwidth]{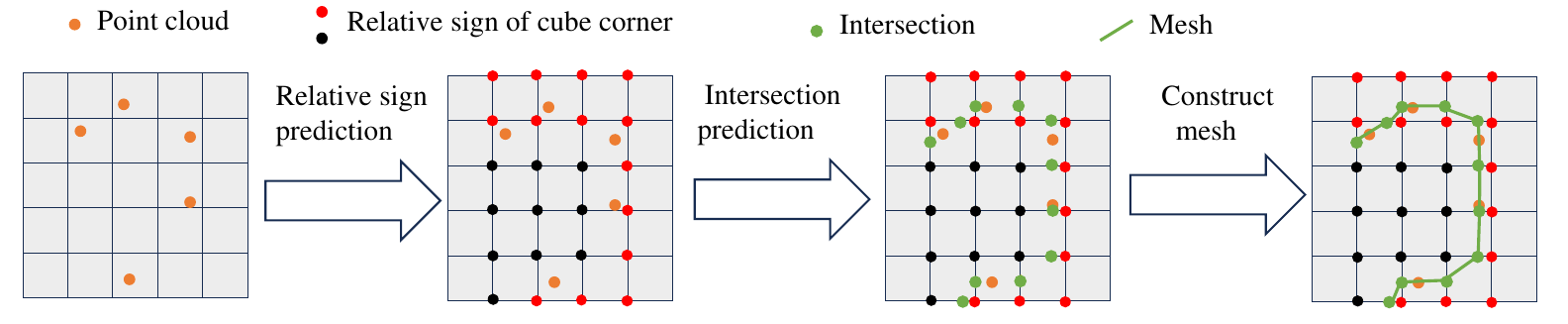}
  \caption{The pipeline of our method. First, we predict the relative sign of the vertices at the corners of the cube. Then, we predict the intersection between the line segment of the point pair and the implicit surface. Finally, we construct the mesh using the template in Marching Cubes.}
  \label{fig:pipe}
\end{figure*}

In this section, we will initially present the complete pipeline of our method. Subsequently, we will delve into the prediction of intersection points and the pair-based sampling technique. Following that, we will discuss the two distinct components: predicting the relative sign and the intersection point individually. Lastly, we will elucidate the training loss.

\subsection{Architecture}

The total pipeline is shown in Fig.~\ref{fig:pipe}. 
The network architecture is depicted in a straightforward manner, as illustrated in Fig.~\ref{fig:network}. Initially, the input point cloud with a total of $N$ points is inputted into the encoder to extract features with a dimension of $N*256$. The encoder comprises 4 transformer layers\cite{zhao2021pointtrans, 2020Vector}, outputting dimensions of $64$, $256$, $256$, and $256$ respectively. Following a similar approach to GIFS~\cite{gifs}, we adopt a paired query point sampling method, generating a total of $M$ pairs. Subsequently, an interpolation layer is employed to derive the features of the $2M$ query points. The network then diverges into two distinct paths. In the upper path, the focus is on predicting the relative sign of the two points. The feature of the point pair is fed into the Relative Sign Module to capture the relationship between the points. A multi-layer perceptron (MLP) is utilized to predict whether the two points exhibit the same sign, with a binary cross-entropy loss function applied. In the bottom path, the aim is to predict the position of the intersection point. The feature of the point pair is inputted into the Intersection Module, followed by the use of a MLP to predict the relative distance from the starting point to the intersection point. A simple regression loss is employed for this prediction. However, in cases where no intersection occurs between the cube edge and the surface, this term is disregarded in the loss calculation.

\subsection{Pair Prediction}
To sample point pairs for training and testing, we start by considering a point cloud $\mathcal{P}=\{p_1, p_2,\cdots, p_N\}$. During training, we uniformly sample points around $\mathcal{P}$ at varying scales and ratios. For each point $p_i$, a query point $q=p_i + U(-1, 1)\times s$ is randomly sampled, where $s$ takes values of 0.005, 0.01, and 0.02, corresponding to sample ratios of 0.6, 0.3, and 0.1 respectively. Subsequently, a cube with an edge length of 1/256 and $q$ as the upper-left-front corner is generated, allowing us to sample 12 point pairs corresponding to the 12 edges of the cube. Following this, we determine the intersection point of each edge with the surface and assess whether the point pairs lie on the same side of the surface.

\begin{figure*} 
\centering
  \includegraphics[width=1.8\columnwidth]{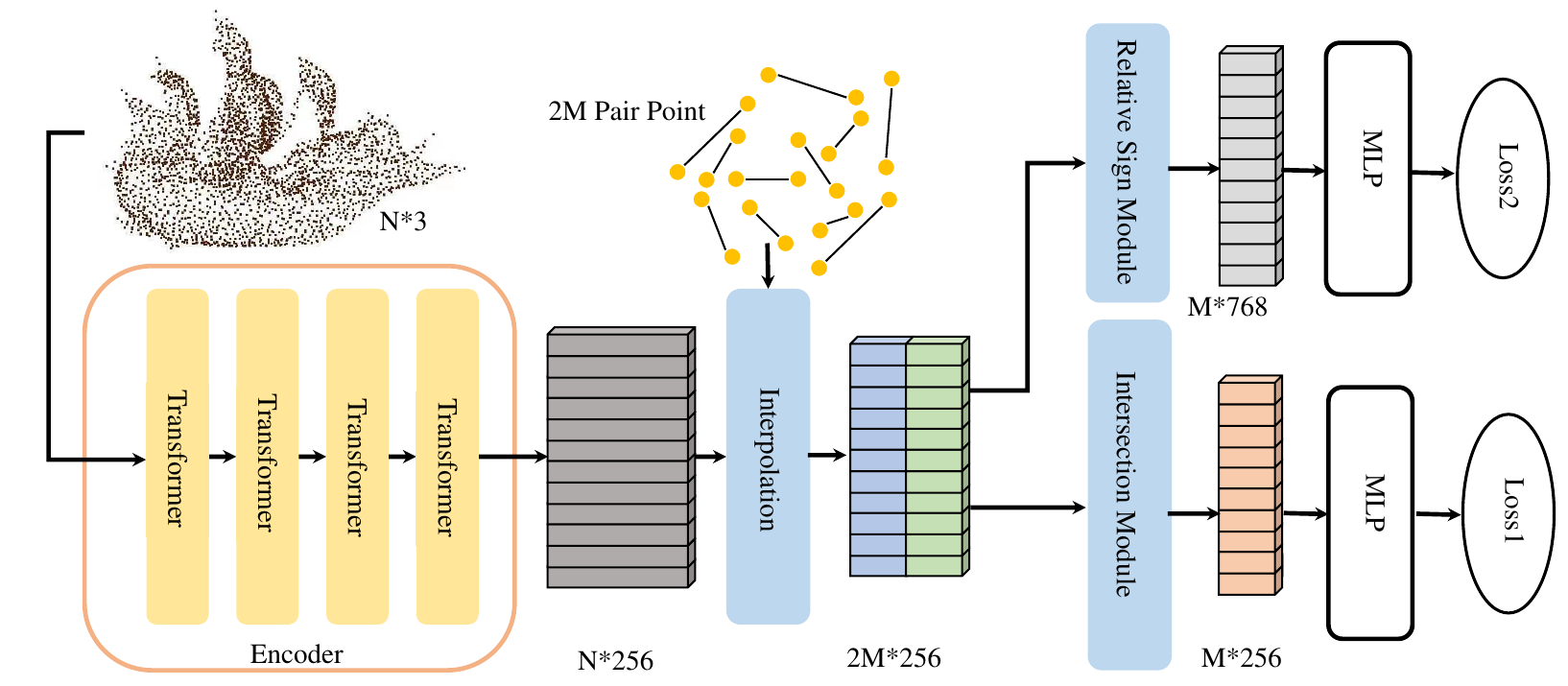}
  \caption{The network architecture of our method, where $M$ is the number of point pair, $N$ is the number of points in a point cloud.}
  \label{fig:network}
\end{figure*}

During testing, to reconstruct the entire surface within the given space, we begin by identifying the bounding box of the shape and dividing it into cube meshes with an edge length of 1/256. For each cube within the bounding box, we predict the relative sign of the cube's corners and the intersection positions, where the reference point is the vertex at the upper-left-front corner of the cube. Utilizing the template matching method in Marching Cubes, we can then generate the triangle mesh.
After sampling, we will elaborate on the specific predictions made by our network. Illustrated in Fig.~\ref{fig:network}, the lower pathway involves selecting a point pair AB, with point $A$ designated as the starting point and $B$ as the end point. The intersection point is denoted as $P$, and our aim is to forecast the proportionate distance from $A$ to $P$, expressed as $\frac{|AP|}{|AB|}$. Furthermore, as depicted in Fig.~\ref{fig:network}, within the upper pathway, we predict whether the point pairs lie on the same or different sides of the surface. The Relative Sign Module's name suggests that we solely predict the relationship between the corners of the cube and the upper-left-front corner within the cube. It is not the sign in the global view but a relative sign to the upper-left-front corner.
\subsection{Relative Sign Module}
In the Relative Sign Module, we formalize the problem as follows: given a start point $A$ and an end point $B$ with features $f_A$ and $f_B$ respectively, we denote the Relative Sign Module as $G(f_A, f_B)$. The Relative Sign Module $G(f_A, f_B)$ is required to exhibit the following property: when the positions of $A$ and $B$ are exchanged, $G(f_A, f_B)$ should vanish, i.e., $G(f_A, f_B) = G(f_B, f_A)$. This is because the relationship between $A$ and $B$ remains unchanged when their positions are swapped. To maintain this property and enhance the expressive capacity of $G(\cdot,\cdot)$, we design the Relative Sign Module as follows:
\begin{equation}
    G(f_A, f_B) = concat(f_A+f_B+pos, f_A*f_B+pos, max(f_A, f_B)+pos),
\end{equation}
where $pos$ is the positional encoding\cite{2020Fourier} of $A-B$. Here, in order to keep the property $G(f_A, f_B)=G(f_B, f_A)$, we abandon the traditional sin-cos positional encoding and adopt only cosine positional encoding. Specifically, $pos(A, B)=cos(\frac{A-B}{2^i})$, $i \in \{0,2,\cdots,19\}$. 

\subsection{Intersection Module}
In the Intersection Module, we aim to predict the position of the intersection point $P$ between point pair $A$ and $B$, each associated with features $f_A$ and $f_B$ respectively. The position of $P$ is represented by $\alpha \in [0, 1]$, where $\alpha=\frac{|PA|}{|AB|}$. To ensure the continuity of the surface, we define the Intersection Module as $H(f_A, f_B)$, which should satisfy the property $H(f_A, f_B)=1-H(f_B, f_A)$. This condition is crucial for maintaining surface continuity. The relationship can be intuitively understood by observing that $\frac{|PA|}{|AB|} = 1-\frac{|PB|}{|AB|}$. Based on this, we formulate the design of the Intersection Module as follows,
\begin{equation}
    H(f_A, f_B) = sigmoid(f_A-f_B+pos_2),
\end{equation}
where $pos_2$ is also the positional encoding of $A-B$. However, in this part, in order to keep the property $H(f_A, f_B)=1-H(f_B, f_A)$, we adopt only sine positional encoding. Specifically, $pos_2(A, B)=sin(\frac{A-B}{2^i})$, $i \in \{0,2,\cdots,19\}$. 

\begin{figure} 
\centering
  \includegraphics[width=0.9\columnwidth]{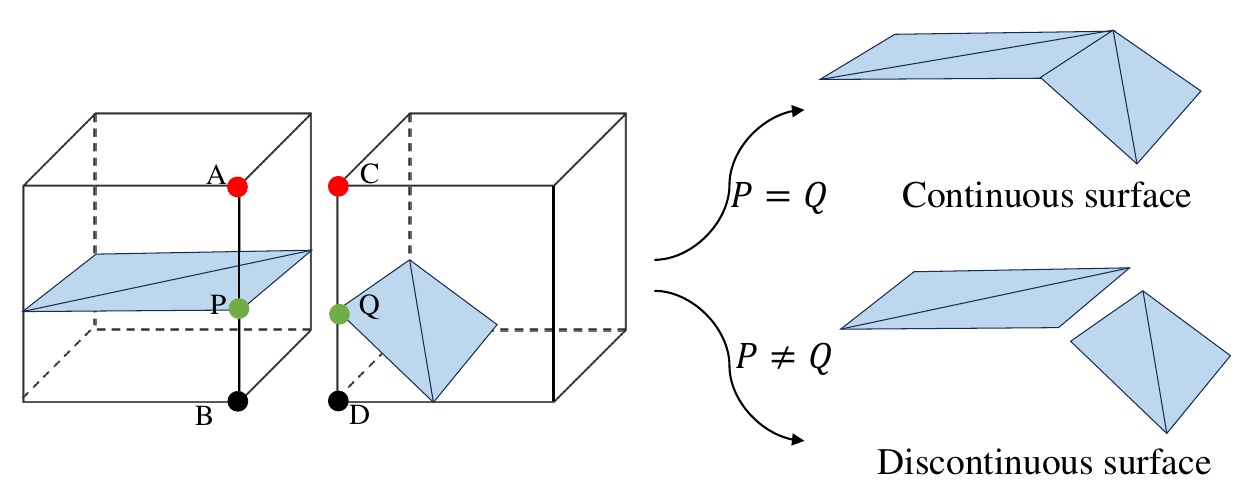}
  \caption{The symmetry ensures the continuity of surface.}
  \label{fig:connection}
\end{figure}

Next, we can delve into the explanation for why a surface may not exhibit continuity if $H(f_A, f_B)\neq 1-H(f_B, f_A)$. Referring to Fig.~\ref{fig:connection}, the two cubes represent adjacent cubes in the Marching Cube algorithm. It is evident that $A$ is equivalent to $C$ and $B$ is equivalent to $D$. Pints $P$ and $Q$ must coincide; otherwise, the mesh will lack continuity. Therefore, the relationship $\frac{|AP|}{|AB|}=1-\frac{|DQ|}{|CD|}$ must be upheld. This condition serves as both necessary and sufficient for ensuring $H(f_A, f_B)=1-H(f_B, f_A)$ and $G(f_A, f_B) = G(f_B, f_A)$.

\subsection{Losses}
In this approach, we utilize two types of loss functions for training. For sign prediction, we employ a straightforward binary cross-entropy loss. For intersection prediction, we apply a regression loss to minimize the disparity between the predicted $\alpha_j$ and the ground-truth $\alpha_{j_{GT}}$, as illustrated below:
\begin{equation}
    Loss_1 = \sum_{j=1}^{M}[\alpha_j - \alpha_{j_{GT}}]_+,
\end{equation}
where $\alpha_j$ means the predicted intersection position of $j-th$ point pair, $\alpha_{j_{GT}}$ means the ground-truth intersection position of $j-th$ point pair,
\begin{equation}
[\alpha-\alpha_{GT}]_{+}= 
\left\{ 
    \begin{array}{c}
        |\alpha-\alpha_{GT}|, \alpha_{GT} \in [0, 1] \\
        0,\alpha_{GT} \notin [0, 1].\\
    \end{array}
\right.
\end{equation}

\subsection{Details and Discussions}

\paragraph{Training.} During training, we utilize 16,000 randomly sampled point pairs every iteration. The training epoch is 100. We employ the Adam optimizer~\cite{adam} and incorporate cosine learning rate adjustment~\cite{2016coslr}. The initial learning rate is set to 0.001, and the batch size is 1.
\paragraph{Testing.} During testing, to assess the quality of the reconstructed mesh, we randomly sample 100,000 points with normals on both the reconstructed mesh and the ground-truth mesh. Subsequently, we calculate the Chamfer Distance ($CD_1$) and Normal Consistency ($NC$) between these point sets.

\paragraph{Training/Testing splitting.} During training, we sampled 1300 shapes for ShapeNet, 71 for MGN, and 100 for ScanNet, respectively. For testing, the same number of shapes was sampled. It is important to note that the shapes used for training and testing do not overlap.

\paragraph{Two prediction ways.}
In fact, we have experimented with two methods for constructing the mesh. The first method involves first predicting the relative sign of the cube corners and then predicting the intersection point when the point pair is on different sides of the surface, with the intersection parameter $\alpha \in [0, 1]$. On the other hand, the second method directly predicts the intersection parameter $\alpha \in \mathbb{R}$ without the need to predict the cube corner sign first. By observing that if $\alpha \in [0, 1]$, an intersection exists between the two points, while no intersection point exists for other values of $\alpha$, we found that the convergence of the second method was unsatisfactory in our experiments. Therefore, we have decided to adopt the first method for constructing the mesh.

\paragraph{Post-processing.} After generating the mesh, we design a method to remove the triangle soup which is obviously wrong. The idea is very simple. In short, if the normal of the triangle is far from its neighbors, we will remove this triangle. Specifically, we calculate the following indicator 
\begin{equation}
    I = \frac{1}{k}\sum_{i=0}^{k-1}|<n, n_i>|,
\end{equation}
where $k$ is number of k-nearest neighbor, $<\cdot, \cdot>$ is inner-product. $n$ is the normal of triangle, $n_i$ is the normal of its neighbor triangles. If $I<0.65$, we will remove the triangle. 

\paragraph{Similarities and differences between GIFS~\cite{gifs} and Ours.}
Our approach shares a common point with GIFS in that both methods involve predicting the relative sign of a point pair, indicating whether the two points are on the same side or opposite sides of the surface. However, a key distinction between our method and GIFS lies in the approach to determining the intersection point: we directly predict the intersection between a segment and a surface, whereas GIFS utilizes $\alpha=\frac{UDF_1}{UDF_2}=\frac{|PA|}{|AB|}$ to define point $P$. Additionally, a minor difference is observed in the network architecture, as we employ a point transformer and interpolation layer to extract query point features, while GIFS utilizes 3D convolution for the same purpose.

\paragraph{Similarities and differences of Deep Marching Cube, Neural Marching Cube and Ours.}
Deep Marching Cube~\cite{2018Deepmc} is a work closely related to ours, as both focus on the two key aspects: the sign of vertices at the corner of the cube and the position of intersection points. However, several differences distinguish our approaches. Firstly, our motivation differs. While our goal is to eliminate artefacts, Deep Marching Cube aims to differentiate the Marching Cube. Secondly, our method is UDF-based, allowing us to represent open surfaces, whereas Deep Marching Cube, based on SDF, can only represent watertight surfaces. Thirdly, we directly predict the relative sign of point pairs, while Deep Marching Cube employs a probabilistic model to represent vertex signs.

Neural Marching Cube~\cite{neuralmc} seeks to enhance Marching Cube performance by intersecting additional vertices beyond those on the cube edges, aiming to generate a finer mesh without cube subdivision. Consequently, our approach and Neural Marching Cube employ different methods to address distinct problems, indicating a somewhat distant relationship between them.

\section{Experiments}

\subsection{Tiny example to show the difficulty of learning UDF}\label{sec:hard_udf}
\begin{figure}
\centering
  \includegraphics[width=1\columnwidth]{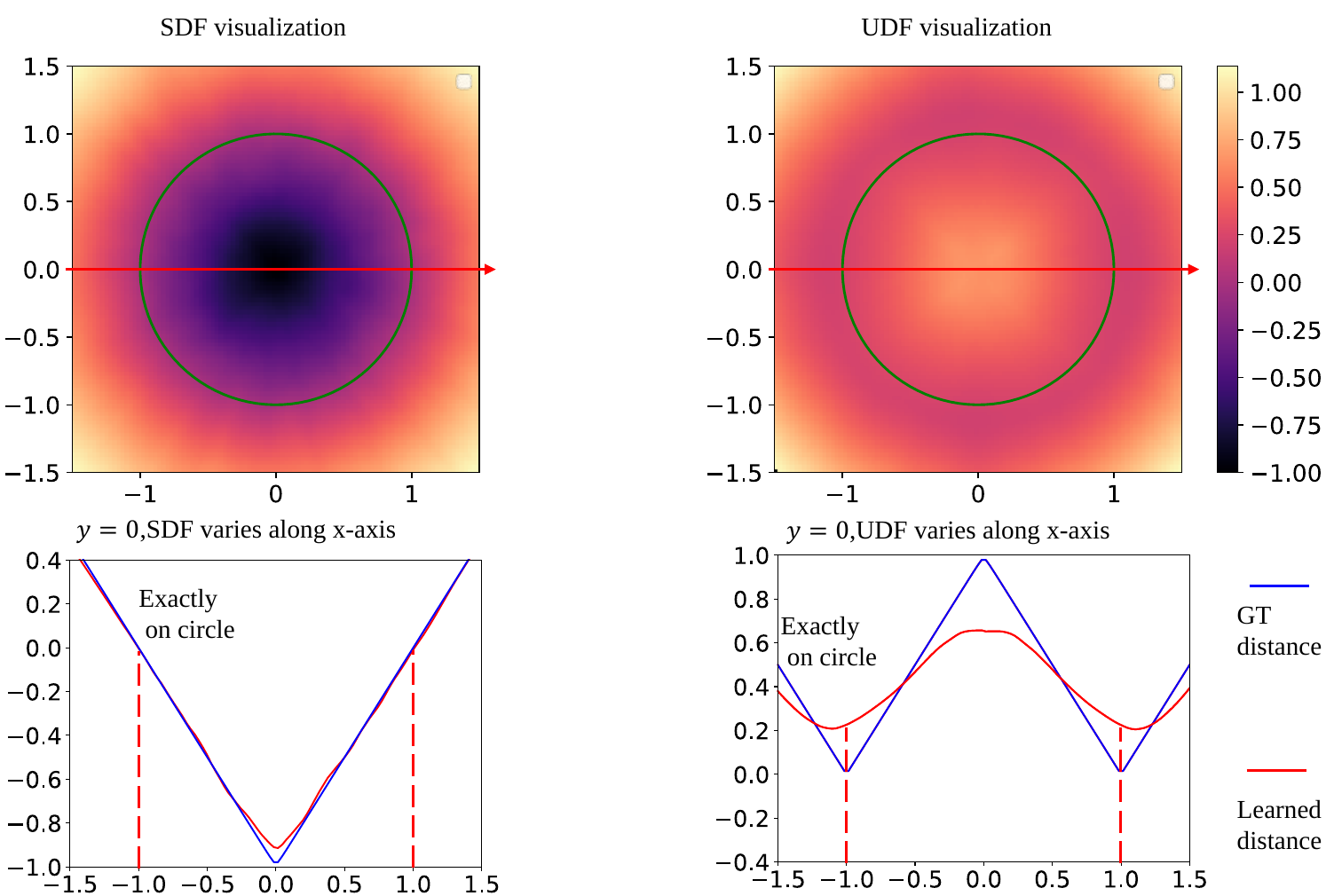}
  \caption{Comparison of learning difficulties between UDF and SDF. Using neural networks to learn the Signed Distance Function (SDF) and Unsigned Distance Function (UDF) from any point in the plane to the circle represented by the green line in the first row of the image. The second row shows the variation of SDF and UDF along the x-axis from the position at $y=0$.}
  \label{fig:error}
\end{figure}

Compared to SDF, UDF does not require predicting signs, making it seem simpler than SDF. However, this is not the case. We designed a small experiment to demonstrate the difficulties of learning UDF, using a simple neural network to fit the circle function on a two-dimensional plane. Our goal is to learn the UDF and SDF of a circle on the plane, using a three-layer MLP (Multi-Layer Perception) to represent this sphere, with optimization done through gradient descent. We provide two three-layer MLP functions $\mathcal{F}_{\theta_1}:\mathbb{R}^3 \rightarrow \mathbb{R}^+$ to represent the UDF field on the plane, and $\mathcal{F}_{\theta_2}:\mathbb{R}^3 \rightarrow \mathbb{R}$ to represent the SDF field on the plane. The parameters of the experiment are as follows: the output dimensions of the three-layer MLP are $32, 64, 1$ respectively, the number of sampled points is $20000$, the initial learning rate is $0.001$, and the optimizer used is Adam. The visualization of the learning results is shown in Fig.~\ref{fig:error}. 

From the results of SDF and UDF in the first row, it can be seen that the neural network indeed learns the approximate distance field. When the distance field changes along the x-axis at the position of $y=0$ in the second row, we focus on the error between the GT distance field and the learned distance field to the circle as x changes. It can be observed that when learning SDF, the error between the two is almost zero, while when learning UDF, the error is significant. This is because near the surface of the curve, the gradient of the SDF remains unchanged, resulting in a continuous change in the distance field, while the gradient of the UDF rotates by $180^\circ$, making the change in the distance field discontinuous. Specifically, when x approaches $-1$ from the left side, the gradient of the SDF is
\begin{equation}
    \lim_{x\rightarrow 0^+}\nabla\mathcal{F}_{\theta_2} = (-1, 0)
\end{equation}
When x approaches -1 from the right, the gradient of the SDF is
\begin{equation}
    \lim_{x\rightarrow 0^-}\nabla\mathcal{F}_{\theta_2} = (-1, 0)
\end{equation}
They are same with each other in SDF.
when x approaches $-1$ from the left side, the gradient of the UDF is
\begin{equation}
    \lim_{x\rightarrow 0^+}\nabla\mathcal{F}_{\theta_1} = (-1, 0)
\end{equation}
When x approaches -1 from the right, the gradient of the UDF is
\begin{equation}
    \lim_{x\rightarrow 0^-}\nabla\mathcal{F}_{\theta_1} = (1, 0)
\end{equation}
The two are exactly opposite. Although the Signed Distance Function (SDF) also has some errors at $x=0$ due to the collision of gradient directions, it is insignificant because it occurs at the center of the sphere, which is not important for surface reconstruction. Therefore, the error here can be ignored. However, the Unsigned Distance Function (UDF) has a discontinuity in the vicinity of the surface, and the distance field near the surface is particularly important for surface reconstruction. For regression methods, fitting a continuously changing function is relatively easy, while fitting a function with gradient discontinuities is very difficult. This is the difficulty of reconstruction based on UDF. The difficulty of reconstruction based on SDF is easier to understand, mainly in complex topological surfaces where it is difficult for neural networks to determine the sign information at a specific point.

\subsection{Dataset and Experiment Settings}

\begin{figure*} 
\centering
  \includegraphics[width=2.1\columnwidth]{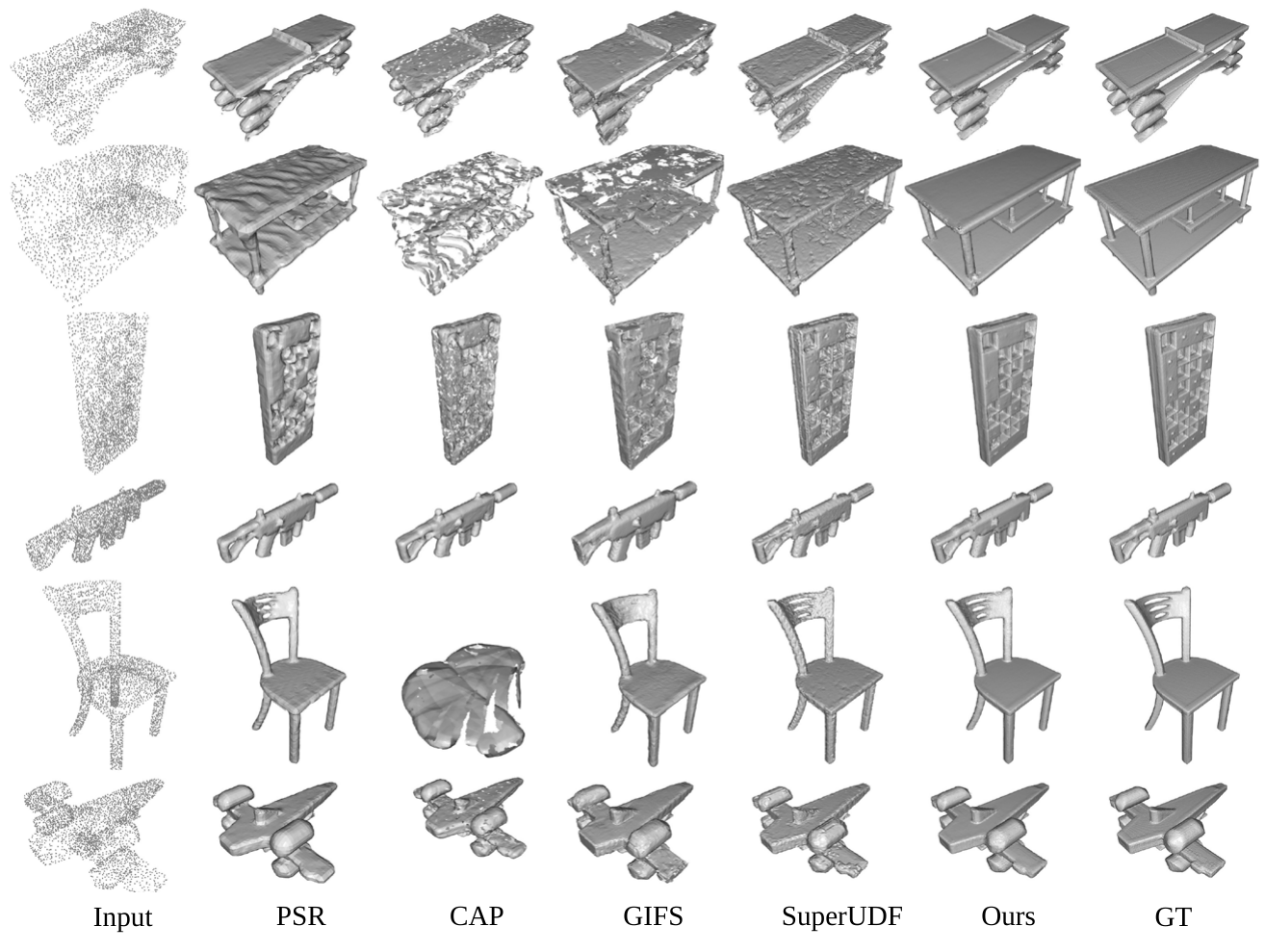}
  \caption{Visualized results of our method and state-of-the-art alternatives. Note that we provide the ground-truth normal as input for PSR as it requires. Methods are PSR\cite{2013Poisson}, CAP\cite{capudf}, GIFS\cite{gifs}, SuperUDF\cite{superudf}. The point number of input point cloud is 3000.}
  \label{fig:shapenet_result1}
\end{figure*}
Our methodology is assessed on three prominent open-source datasets. Initially, we randomly select 1300 shapes from 13 distinct categories within the ShapeNet dataset~\cite{chang2015shapenet}, with each category comprising 100 shapes. The point number of each input point cloud is 3000. Subsequently, we employ the MGN dataset, encompassing both pants and shirts. The point number of each input point cloud is 3000. Finally, we include 100 scenes from the real-world ScanNet dataset~\cite{dai2017scannet}, obtained through scanning technology, resulting in 10,000 input point clouds. To facilitate a comprehensive comparison with existing techniques, we utilize standard evaluation metrics commonly employed in surface reconstruction. One such metric is Chamfer-$L_1$ ($CD_1$), which measures the distance between two sets of points. Additionally, to demonstrate enhancements in artifact elimination, we utilize normal consistency ($NC$) as a metric to evaluate the smoothness of the reconstructed surface.

When calculating $CD_1$, we follow O-Net\cite{2019Occupancy} and randomly sample 100000 points on the reconstructed mesh and 100000 points on ground-truth mesh. $CD_1$ equation is shown as the Eq.\ref{eqn:chamfer}
\begin{equation}
\label{eqn:chamfer}
\begin{aligned}
\mathrm{CD}_{1}=& \frac{1}{2 N_{x}} \sum_{i=1}^{N_{x}}\left\|\mathbf{x}_{i}-\mathcal{S}_{y }\left(\mathbf{x}_{i}\right)\right\|_1+\\
& \frac{1}{2 N_{y}} \sum_{i=1}^{N_{y}}\left\|\mathbf{y}_{i}-\mathcal{S}_{x}\left(\mathbf{y}_{i}\right)\right\|_1,
\end{aligned}
\end{equation}
$\{ \mathbf{x}_{i}, i=0\cdots N_x\}$ is the points set sampled from ground-truth mesh. $\{\mathbf{y}_{i}, i=0\cdots N_y\}$ is the point set sampled from reconstructed mesh. $\mathcal{S}_{y}(\mathbf{x}_{i})$ means the nearest point to $\mathbf{x}_{i}$ in reconstructed point set. $\mathcal{S}_{x}(\mathbf{y}_{i})$ means the nearest point to $\mathbf{y}_{i}$ in ground-truth point set. $\|\cdot\|_1$ means $L1-distance$.

Normal consistency is closely related to $CD_1$, shown in Eq.\ref{eqn:nc}
\begin{equation}
\label{eqn:nc}
\begin{aligned}
\mathrm{NC}=& \frac{1}{2 N_{x}} \sum_{i=1}^{N_{x}}|<\mathbf{n}\left(\mathbf{x}_{i}) , \mathbf{n}(\mathcal{S}_{y}\left(\mathbf{x}_{i}\right)\right)>|+\\
& \frac{1}{2 N_{y}} \sum_{i=1}^{N_{y}}|<\mathbf{n}\left(\mathbf{y}_{i}) , \mathbf{n}(\mathcal{S}_{x}\left(\mathbf{y}_{i}\right)\right)>| .
\end{aligned}
\end{equation}
$\mathbf{n}(x)$ means the normal of point $x$. $<,>$ means inner-product. Here, we calculate the absolute value of inner product, because our mesh is not orientated. 

\subsection{Watertight Surface Reconstruction}
\begin{figure*} 
\centering
  \includegraphics[width=2.1\columnwidth]{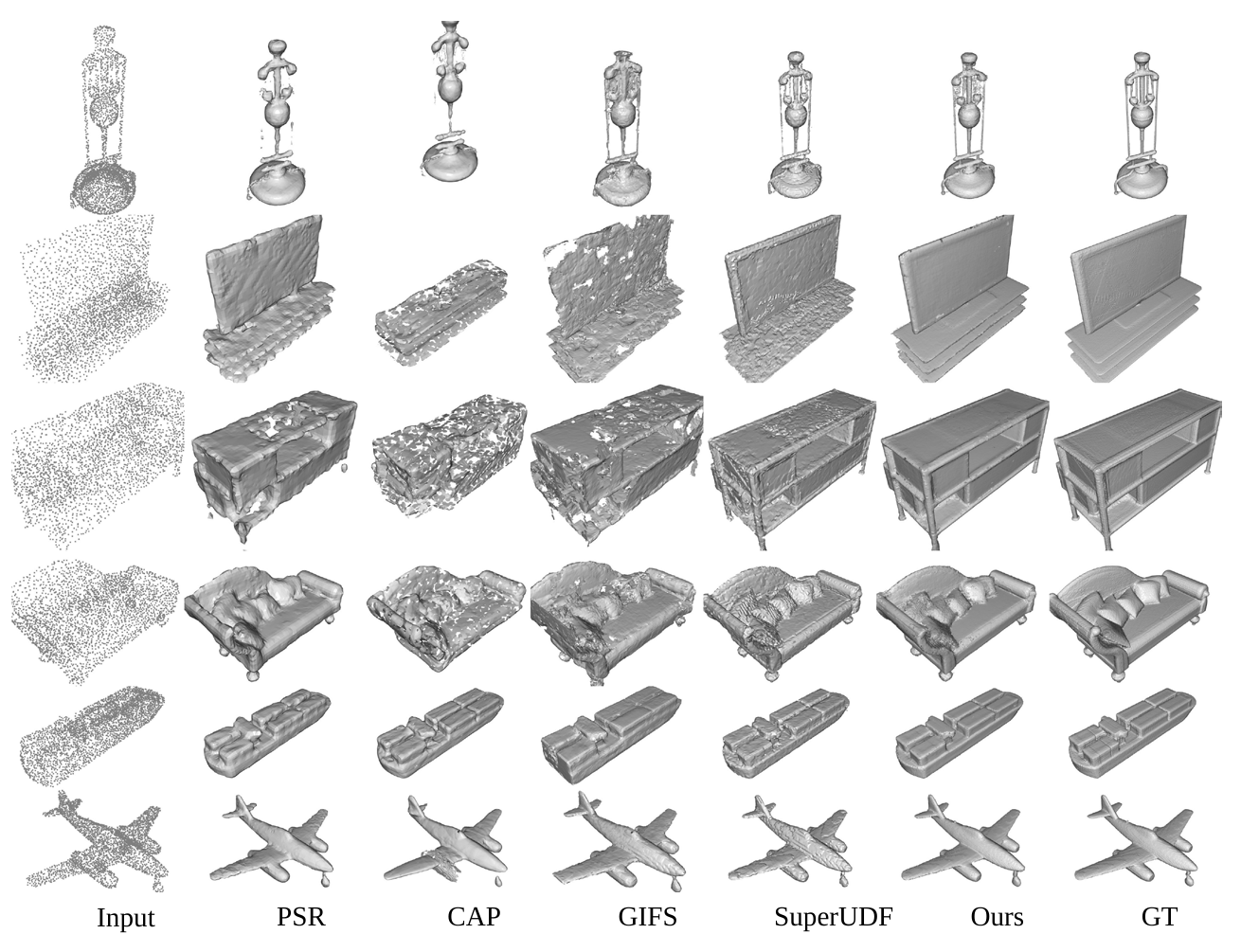}
  \caption{Visualized results of our method and state-of-the-art alternatives. Note that we provide the ground-truth normal as input for PSR as it requires. Methods are PSR\cite{2013Poisson}, CAP\cite{capudf}, GIFS\cite{gifs}, SuperUDF\cite{superudf}. The point number of input point cloud is 3000.}
  \label{fig:shapenet_result2}
\end{figure*}

\begin{table*}[]
    \centering
\begin{tabular}{l|lllll|lllll}
\toprule
Indicator&\multicolumn{5}{c}{\text { $CD_1 \downarrow$ }} &\multicolumn{5}{c}{\text { $NC \uparrow$ }}\\
\midrule
Class     & GIFS            & CAP    & PSR & SuperUDF & Ours&   GIFS            & CAP    & PSR & SuperUDF & Ours            \\
\midrule
airplane  & 0.0026          & 0.0031 & 0.0023  & 0.0022 & \textbf{0.0012} & 0.9322          & 0.9318 & 0.9435  & 0.9469 & \textbf{0.9716}\\
bench     & 0.0058          & 0.0059 & 0.0037  & 0.0029&\textbf{0.0019}& 0.8865          & 0.8874 & 0.9062  & 0.9234&\textbf{0.9449} \\
cabinet   & 0.0104     & 0.0048 &0.0056 & 0.0039&\textbf{0.0026}& 0.9370          & 0.9331 & 0.9357  & 0.9456&\textbf{0.9700} \\
car       & 0.0052          & 0.0054 & 0.0039   & 0.0029&\textbf{0.0024}& 0.8825          & 0.8879 & 0.8975  & 0.9014&\textbf{0.9133} \\
chair     & 0.0041          & 0.0042 & 0.0053 &0.0039& \textbf{0.0024} & 0.9415          & 0.9447 & 0.9311  & 0.9498&\textbf{0.9601}\\
display   & 0.0055          & 0.0047 & 0.0043  & 0.0034& \textbf{0.0021}& 0.9456          & 0.9513 & 0.9643  & 0.9712&\textbf{0.9834} \\
lamp      & 0.0078          & 0.0082 & 0.0036 &0.0030  & \textbf{0.0018}& 0.9108          & 0.9025 & 0.9366  & 0.9334&\textbf{0.9476} \\
speaker   & 0.0081          & 0.0063 &0.0050  &0.0041  & \textbf{0.0026}& 0.9015          & 0.9243 & 0.9531  & 0.9589&\textbf{0.9712} \\
rifle     & 0.0029 & 0.0013 & 0.0012  & 0.0011  & \textbf{0.0009} & 0.9345          & 0.9533 & 0.9685  & 0.9668&\textbf{0.9776}\\
sofa      & 0.0069 & 0.0050 & 0.0038 & 0.0033&\textbf{0.0022} & 0.9370          & 0.9413 & 0.9534  & 0.9520&\textbf{0.9758}\\
table     & 0.0065          & 0.0088 & 0.0043  & 0.0025&\textbf{0.0023} & 0.9235          & 0.9109 & 0.9381  & 0.9428&\textbf{0.9749}\\
phone & 0.0049          & 0.0024 & 0.0024& 0.0021&\textbf{0.0017} & 0.9537          & 0.9725 & 0.9790  & 0.9713&\textbf{0.9869}\\
vessel    & 0.0033          & 0.0025 & 0.0012  & 0.0021&\textbf{0.0015} & 0.9167          & 0.9355 & 0.9464  & 0.9487&\textbf{0.9625}\\
\midrule
Mean      & 0.0057          & 0.0048 & 0.0037  & 0.0028&\textbf{0.0019}& 0.9239          & 0.9284 & 0.9423  & 0.9470&\textbf{0.9648}\\
\bottomrule
\end{tabular}
\vspace{3pt}
    \caption{$CD_1$ and $NC$ comparison on ShapeNet within 13 classes. $\downarrow$ means lower is better. $\uparrow$ means higher is better. Comparing methods are  GIFS~\cite{gifs},CAP~\cite{capudf},PSR~\cite{2013Poisson},SuperUDF~\cite{superudf}. The point number of input point cloud is 3000.}
    \label{tab:shapenet}
\end{table*}
Several previous studies~\cite{2019DeepSDF, 2020Convolutional, 2019Occupancy, 2020ndf} have commonly utilized the watertight surface reconstruction approach for implicit surface reconstruction, a key focus of this point cloud reconstruction research. In order to evaluate the effectiveness of our proposed method, we conducted a comprehensive analysis of watertight surface reconstruction using the ShapeNet dataset~\cite{chang2015shapenet}. Specifically, we uniformly sampled 3000 points from the watertight mesh to serve as input data. We compared our method against four well-known techniques, excluding PSR~\cite{2013Poisson}, all of which are UDF-based methods. The methods included GIFS~\cite{gifs}, CAP~\cite{capudf}, and SuperUDF~\cite{superudf}. For the PSR~\cite{2013Poisson} method, we incorporated ground-truth normals as input data. The visualization results are presented in Fig.~\ref{fig:shapenet_result1} and Fig.~ \ref{fig:shapenet_result2}. Our method not only preserves more details but also produces smoother results. One of the reasons for the unsatisfactory performance of CAP \cite{capudf} is the insufficient number of input points in the point cloud. With only 3000 points available for training, CAP typically requires over 20000 points in the point cloud to achieve optimal results. Typically, methods such as ConvOccNet\cite{2020Convolutional} and POCO\cite{2022POCO} initially learn features from sparse point clouds and then decode the distance field of query points. Conversely, approaches like Neural Pull\cite{2020npull} and CAP\cite{capudf} train neural networks to approximate the distance field, often necessitating dense point clouds. Our method falls within the former category.

The quantitative evaluation results are summarized in Table~\ref{tab:shapenet}. These results demonstrate that our method achieves excellent performance across all 13 classes with a significant margin, showcasing its state-of-the-art capabilities. Particularly noteworthy is the performance on the $NC$, indicating that our reconstructed surfaces are not only faithful to the ground truth but also exhibit smoothness.

\subsection{Open Surface Reconstruction}
\begin{figure*} 
\centering
  \includegraphics[width=2\columnwidth]{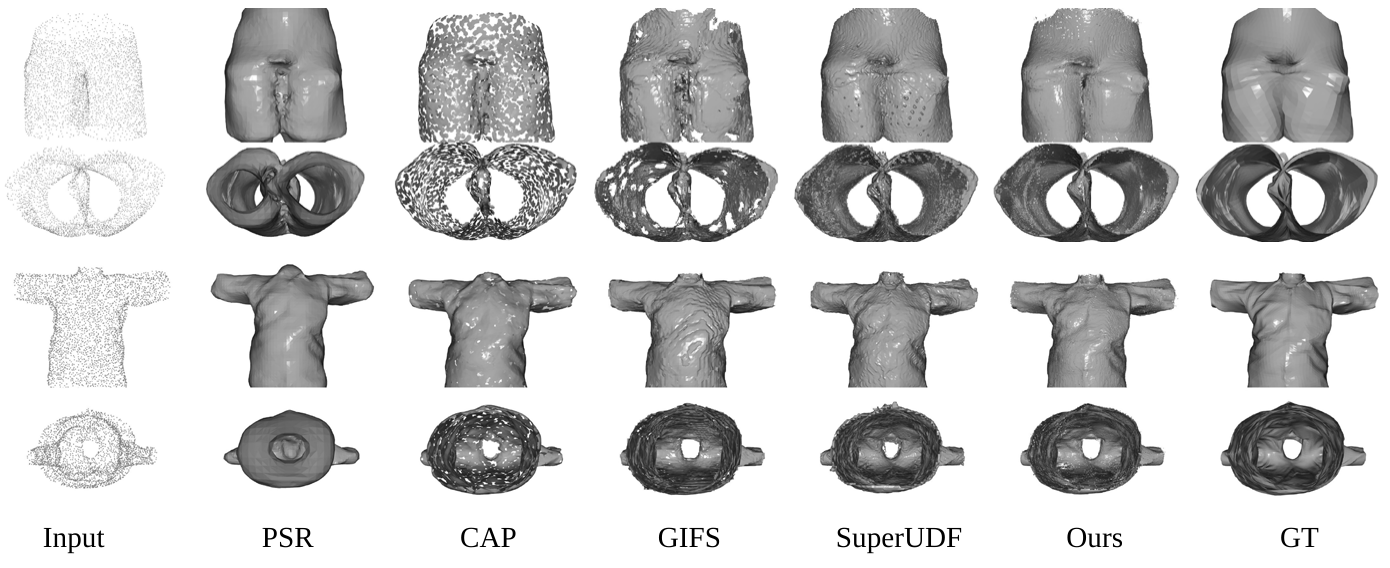}
  \caption{Visualized results of our method and state-of-the-art alternatives. Note that we provide the ground-truth normal as input for PSR as it requires. Methods are PSR\cite{2013Poisson}, CAP\cite{capudf}, GIFS\cite{gifs}, SuperUDF\cite{superudf}. The point number of input point cloud is 3000.}
  \label{fig:mgn_result1}
\end{figure*}
\begin{figure*} 
\centering
  \includegraphics[width=2\columnwidth]{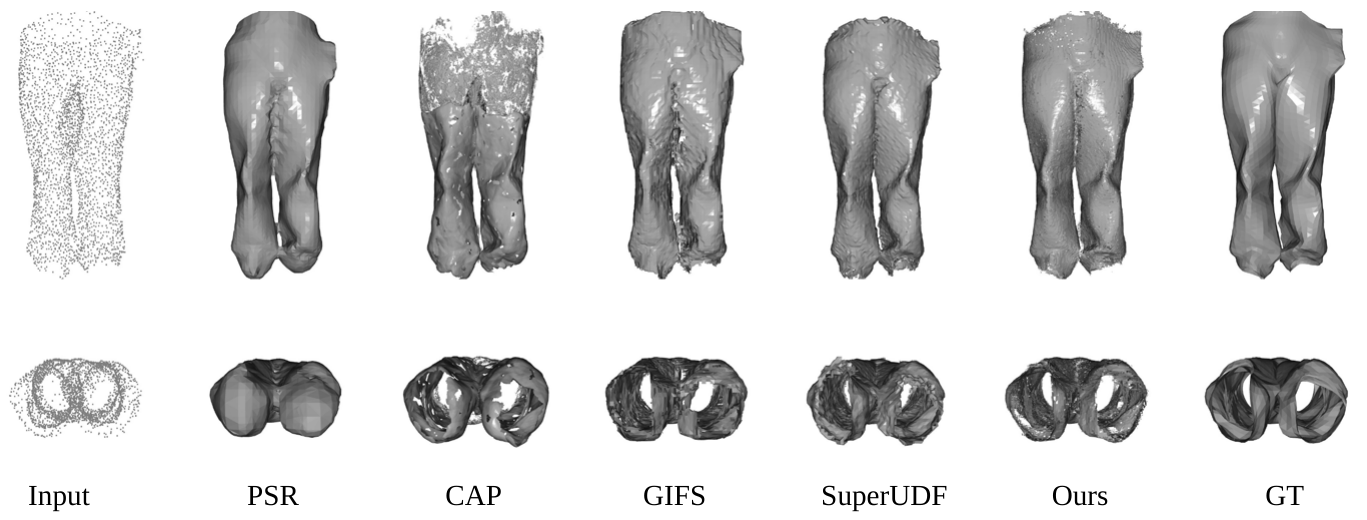}
  \caption{Visualized results of our method and state-of-the-art alternatives. Note that we provide the ground-truth normal as input for PSR as it requires. Methods are PSR\cite{2013Poisson}, CAP\cite{capudf}, GIFS\cite{gifs}, SuperUDF\cite{superudf}. The point number of input point cloud is 3000.}
  \label{fig:mgn_result2}
\end{figure*}

\begin{table}[]
\centering
\begin{tabular}{l|lllll}
\toprule
      & CAP      & PSR    & GIFS      & SuperUDF& Ours     \\
\midrule
$CD_1 \downarrow$& 0.0035 & 0.0045 & 0.0033 & 0.0024& \textbf{0.0019}\\
$NC \uparrow$ & 0.9407 & 0.9356&0.9418 & 0.9645 &\textbf{0.9758} \\
\bottomrule
\end{tabular}
\vspace{3pt}
\caption{$CD_1$ and $NC$ comparison results on MGN dataset. Comparing methods are  GIFS~\cite{gifs},CAP~\cite{capudf},PSR~\cite{2013Poisson},SuperUDF~\cite{superudf}. The point number of input point cloud is 3000.}
\label{tab:garment}
\end{table}
By leveraging a pair-point structure akin to GIFS for representing surfaces within our framework, we are able to proficiently reconstruct open surfaces. We achieve this by uniformly sampling 3000 points on each MGN mesh to reconstruct the implicit surface, a task that poses challenges for SDF-based methods given the characteristics of open surfaces. The comparative analysis presented in Table~\ref{tab:garment} illustrates the superior performance of our approach in comparison to various methods including PSR~\cite{2013Poisson}, CAP~\cite{capudf}, GIFS~\cite{gifs}, and SuperUDF~\cite{superudf}. Moreover, the visual results depicted in Fig.~ \ref{fig:mgn_result1} and Fig.~ \ref{fig:mgn_result2}. exhibit the faithfulness of our reconstructed mesh, capturing intricate details and closely mirroring the ground-truth.

\subsection{Scene Reconstruction}
\begin{figure*} 
\centering
  \includegraphics[width=2\columnwidth]{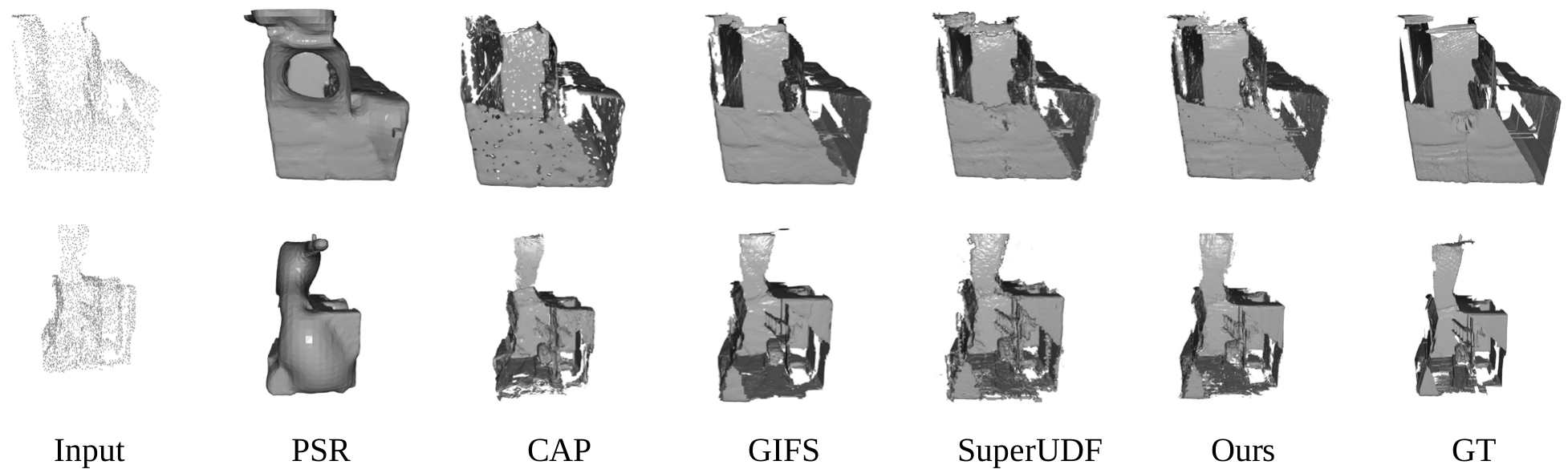}
  \caption{Visualized results of our method and state-of-the-art alternatives. Note that we provide the ground-truth normal as input for PSR as it requires. Methods are PSR\cite{2013Poisson}, CAP\cite{capudf}, GIFS\cite{gifs}, SuperUDF\cite{superudf}. The point number of input point cloud is 10000.}
  \label{fig:scan_result2}
\end{figure*}

\begin{figure*} 
\centering
  \includegraphics[width=2\columnwidth]{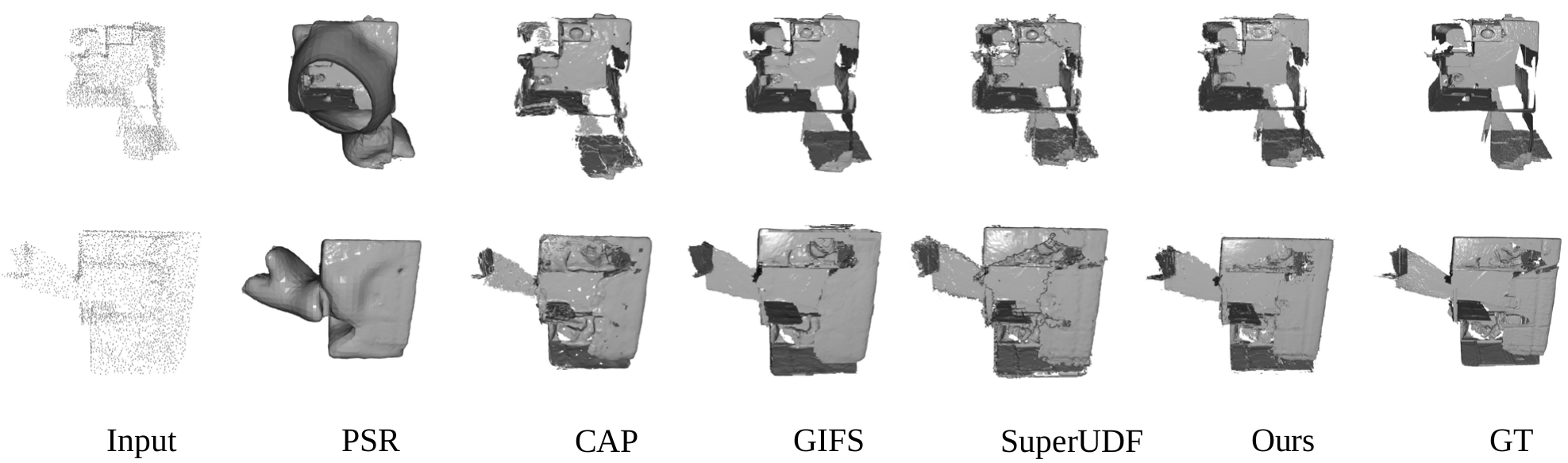}
  \caption{Visualized results of our method and state-of-the-art alternatives. Note that we provide the ground-truth normal as input for PSR as it requires. Methods are PSR\cite{2013Poisson}, CAP\cite{capudf}, GIFS\cite{gifs}, SuperUDF\cite{superudf}. The point number of input point cloud is 10000.}
  \label{fig:scan_result1}
\end{figure*}

Real scene reconstruction is characterized by a heightened level of complexity, primarily due to the inherent openness and incompleteness of the input point cloud data. In this study, we meticulously selected 100 scenes from ScanNet and illustrated the visualization results in Fig.~\ref{fig:scan_result1} and Fig.~\ref{fig:scan_result2}, along with the quantitative results in Table~\ref{tab:scan}, in comparison with other methods.

Upon examining the visualization results, our reconstructed surface exhibits a notably smoother appearance compared to other methods. Furthermore, the quantitative analysis reveals that our approach outperforms other methods in terms of both $CD_1$ and $NC$, underscoring the efficacy of our design in accurately predicting the intersection points between the surface and cube edges.
\begin{table}[]
\centering
\begin{tabular}{l|lllll}
\toprule
       & CAP     & PSR     & GIFS      & SuperUDF & Ours    \\
\midrule
$CD_1 \downarrow$  & 0.0044 & 0.0240 & 0.0043 & 0.0039&\textbf{0.0037} \\
$NC \uparrow$ & 0.8679 & 0.8420 & 0.8713 & 0.8722&\textbf{0.8810}\\
\bottomrule
\end{tabular}
\vspace{3pt}
\caption{$CD_1$ and $NC$ comparison on ScanNet. Comparing methods are GIFS~\cite{gifs},CAP~\cite{capudf},PSR~\cite{2013Poisson},SuperUDF~\cite{superudf}. The point number of input point cloud is 10000.}
\label{tab:scan}
\end{table}

\subsection{Ablation Study}
\paragraph{Contribution of Relative Sign Module and Intersection Module.} To enhance the quality of surface reconstruction, it is essential to predict the sign of cube corners. This allows for the selection of appropriate triangle templates akin to the Marching Cube algorithm. Additionally, accurate intersection point prediction is crucial. By leveraging this information, adjustments can be made to the triangle positions, resulting in smoother surfaces with reduced artefacts. Our approach consists of two key components: the Relative Sign Module and the Intersection Module. Consequently, we aim to conduct experiments to evaluate the impact of each component on surface quality.
In pursuit of this objective, we have devised 9 comparative experiments to demonstrate the contributions of each component, as presented in Table~\ref{tab:ablation_cd} and Table~\ref{tab:ablation_nc}. Upon examining the results row-wise, it is evident that regardless of the chosen sign prediction method, our intersection results outperform those of GIFS \cite{gifs}. Furthermore, the disparity between our intersection prediction outcomes and ground truth (GT) predictions is minimal, indicating the enhanced accuracy of our Intersection Module. 
When analyzing the results column-wise, it is apparent that regardless of the chosen intersection prediction method, our sign prediction results surpass those of GIFS \cite{gifs}. Moreover, the difference between our sign prediction outcomes and GT sign predictions is negligible, highlighting the efficacy of our Relative Sign Module in improving sign prediction accuracy.

\begin{table}[]
\centering
\begin{tabular}{c|ccc}
\toprule
   $CD_1 \downarrow$       & GIFS Intersection & Our Intersection & GT Intersection \\
          \midrule
GIFS Sign & 0.0057            &  0.0042                &          0.0037       \\
Our Sign  &  0.0025                 & 0.0019           &    0.0018             \\
GT Sign   &      0.0023             &  0.0018                & 0.0017             \\
\bottomrule
\end{tabular}
\caption{The contribution of Relative Sign Module and Intersection Module. The $CD_1$ is obtained on ShapeNet 13 classes.}
\label{tab:ablation_cd}

\end{table}

\begin{table}[]
\centering
\begin{tabular}{c|ccc}
\toprule
   $NC \uparrow$       & GIFS Intersection & Our Intersection & GT Intersection \\
          \midrule
GIFS Sign & 0.9239            &   0.9312               &      0.9403           \\
Our Sign  &  0.9435                 & 0.9648           &       0.9703          \\
GT Sign   &  0.9677                 &  0.9801                & 0.9812\\
\bottomrule
\end{tabular}
\caption{The contribution of Relative Sign Module and Intersection Module.  The $NC$ is obtained on ShapeNet 13 classes.}
\label{tab:ablation_nc}

\end{table}
\paragraph{Accuracy of intersection point.}
Except for the final metrics $CD_1$ and $NC$ on the mesh, we also present intermediate results to demonstrate the enhancements made to the Intersection Module. When reconstructing the mesh, both UDF-based methods and ours attempt to compute the intersection point of the surface and the cube edge. The key difference lies in the approach: the UDF-based method determines the intersection position by inversely considering the UDF value between two points, while our method directly predicts the intersection using a neural network. Specifically, we evaluate the distance between the predicted intersection point and the ground-truth intersection point generated by different methods in Table~\ref{tab:distance}. To elaborate, we initially divide the space into cubes with a resolution of 256. Subsequently, for each cube edge intersecting with the implicit surface, we calculate the distance from the predicted intersection to the ground-truth intersection as $d = |p_{\text{predicted}} - p_{\text{gt}}| \times 256$. These results are obtained from the ShapeNet dataset. It is evident that our method significantly enhances the accuracy of the intersection point.
\begin{table}[]
\centering
\begin{tabular}{c|cccc}
\toprule
Method         & GIFS & SuperUDF& CAP & Ours\\
\midrule
Distance $\downarrow$ &  0.0131 & 0.0127 & 0.0095 & \textbf{0.0084}\\
\bottomrule
\end{tabular}
\caption{The distance from predicted intersection to GT intersection on ShapeNet. Comparing methods are GIFS~\cite{gifs},CAP~\cite{capudf}, SuperUDF~\cite{superudf}.}
\label{tab:distance}
\end{table}


\begin{table}
\centering
\begin{tabular}{l|ll}
\toprule
Method         & $CD_1\downarrow$ & $NC\uparrow$\\
\midrule
$\hat{G}(f_A, f_B)$  & 0.0023    & 0.9573    \\
$G(f_A, f_B)$ & \textbf{0.0019}     & \textbf{0.9648}    \\
\bottomrule
\end{tabular}
\caption{The reconstruction result comparison when Relative Sign Module is symmetrical or asymmetrical to the exchange of start and end point. The result is averaged on 13 classes of ShapeNet.}
\label{tab:sign}
\end{table}

\begin{table}
\centering
\begin{tabular}{l|ll}
\toprule
Method         & $CD_1\downarrow$ & $NC\uparrow$\\
\midrule
$\hat{H}(f_A, f_B)$  & 0.0024    & 0.9425    \\
$H(f_A, f_B)$ & \textbf{0.0019}     & \textbf{0.9648}    \\
\bottomrule
\end{tabular}
\vspace{3pt}
\caption{The Comparison when Intersection Module satisfies $H(f_A, f_B)=1-H(f_B, f_A)$ or not. The result is average on 13 classes of ShapeNet.}
\label{tab:intersection}
\end{table}

\paragraph{Relative Sign Module Symmetry.} In our design, the Relative Sign Module remains symmetrical  when the start point and end point exchange their positions, i.e., $G(f_A, f_B)=G(f_B, f_A)$. Here, we aim to quantitatively evaluate the influence of this design feature. Therefore, we compare the reconstruction results when $G(\cdot, \cdot)$ does not maintain this symmetry with the results when $G(\cdot, \cdot)$ does maintain this symmetry. Specifically, when $G(\cdot, \cdot)$ does not maintain the symmetry, we set
\begin{equation}
    \hat{G}(f_A, f_B) = concat(f_A+pos_3, f_B+pos_3),
\end{equation}
where $pos_3$ is the standard positional encoding of both sine and cosine encoding. 
The quantitative result is in Table~\ref{tab:sign}. We can see the symmetry works.

\begin{figure} 
\centering
  \includegraphics[width=\columnwidth]{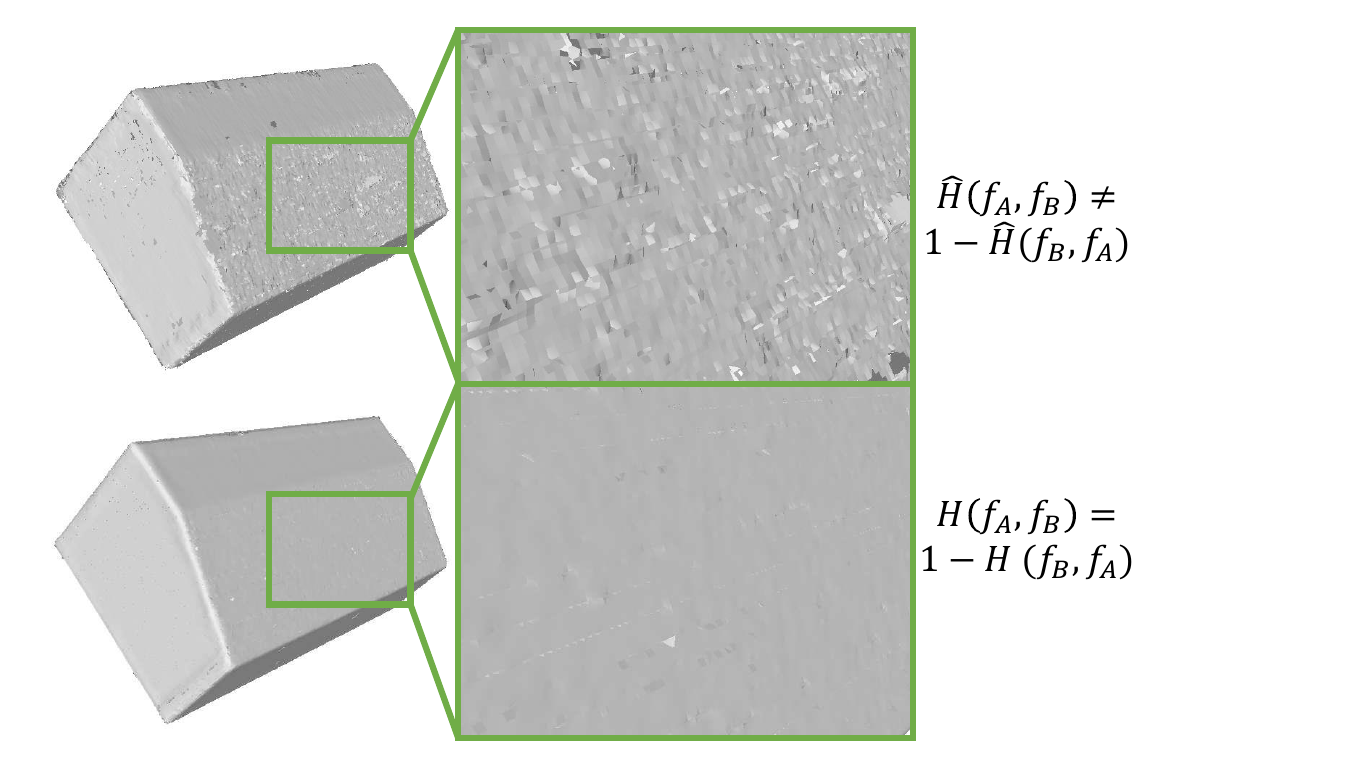}
  \caption{When the Intersection Module satisfies symmetry, the surface is continuous, otherwise it is not continuous and has lots of errors.}
  \label{fig:H}
\end{figure}

\paragraph{Intersection Module symmetry.}
In our design, the Intersection Module exhibits the symmetry of the point pair, i.e., $H(f_A, f_B)=1-H(f_B, f_A)$. This symmetry is ensured by the Sigmoid activation function and the unique design of positional encoding. Here, we seek to assess the influence of this symmetry. Therefore, we compare the results when the symmetry is absent versus when it is present. Specifically, we define the function $\hat{H}(\cdot, \cdot)$ as
\begin{equation}
    \hat{H}(f_A, f_B) = sigmoid(concat(f_A+pos_3, f_B+pos_3)).
\end{equation}
The quantitative result is in Table~\ref{tab:intersection} and visualization result is in Fig.~\ref{fig:H}, we can see that the discontinuity indeed appear when the symmetry is absent. 

\paragraph{Efficiency Analysis}
Here, we present the GPU memory usage and time required for mesh reconstruction in Table~\ref{tab:efficiency}. The average reconstruction time was calculated across 1300 shapes from ShapeNet. The experiments were conducted using a Titan X 12GB GPU and an Intel(R) i9-9900k@3.60GHz CPU. It is evident that our approach achieves high reconstruction speed, albeit at the expense of increased GPU memory utilization.
\begin{table}[]
\centering
\begin{tabular}{c|ccc}
\toprule
Method           & GPU Memory(MB)& Inference Time(s)\\
\midrule
GIFS~\cite{gifs}  & 3478    & 8min46s     \\
CAP~\cite{capudf} & 1398     & 21min56s     \\
Ours & 6756     & 6min52s      \\

\bottomrule
\end{tabular}
\vspace{3pt}
\caption{The time and memory comparison between GIFS~\cite{gifs}, CAP~\cite{capudf} and ours in inference. The result is on the ShapeNet.}
\label{tab:efficiency}
\end{table}

\section{Conclusion and Limitations}
This study introduces a method that predicts the intersection between a line segment of a point pair and an implicit surface to effectively eliminate artefacts. Additionally, it proposes two novel modules that utilize neural networks to predict the relative sign of vertices at the corners of the cube and intersections between surface and line segment. This approach results in a more detailed surface reconstruction with fewer artefacts, leading to excellent outcomes on three datasets: ShapeNet, MGN, and ScanNet.
However, there are some limitations to this method. For example, the method struggles to generate satisfactory meshes near the edges of open surfaces. This is problem will be addressed in the future works.




\bibliographystyle{ACM-Reference-Format}
\bibliography{refbib}



\end{document}